\documentclass{article}

\usepackage{aaai}

\usepackage{latexsym}
\usepackage{amstext,amssymb}

\newcommand{\LPif}{\leftarrow}

\newcommand{\nafo}[0]{\mathit{not}}
\newcommand{\naf}[0]{\nafo\;}

\newcommand{\PREC}[2]{#1\prec #2}

\newcommand{\egc}[0]{e.g.,\ }

\newcommand{\PLP}[0]{\mbox{\sf PLP}}
\newcommand{\DLV}[0]{\texttt{dlv}}
\newcommand{\SM}[0]{\texttt{smodels}}
\newcommand{\ECLiPSe}[0]{{ECLiPSe}}

\title{A Compiler for Ordered Logic Programs\thanks{This work was 
partially supported by the Austrian Science Fund Project N Z29-INF.
}}

\author{
   James P.\ Delgrande\\
     School of Computing Science, \\
     Simon Fraser University, \\
     Burnaby, B.C., \\
     Canada  V5A 1S6, \\
     jim@cs.sfu.ca,
   \And
   Torsten Schaub \\
     Institut f\"ur Informatik, \\
     Universit\"at Potsdam, \\
     Postfach 60 15 53, \\
     D--14415 Potsdam,
     Germany, \\
     torsten@cs.uni-potsdam.de 
   \And
   Hans Tompits \\
     Abt.\ Wissensbasierte Systeme, \\
     Technische Universit\"at Wien, \\
     Favoritenstra{\ss}e~9--11, \\
     A--1040 Wien, Austria, \\
     tompits@kr.tuwien.ac.at
 }

\date{}

\begin{document}

\maketitle

\begin{abstract}
This paper describes a system, called \PLP, for compiling ordered logic 
programs into standard logic programs under the answer set semantics. 
In an ordered logic program, preference information is expressed at the 
object level by atoms of the form $s \prec t$, where $s$ and $t$
are names of rules.
An ordered logic program is transformed into a second, regular, extended 
logic program wherein the preferences are respected,
in that the answer sets obtained in the transformed theory correspond
with the preferred answer sets of the original theory.
Since the result of the translation is an extended logic program, existing
logic programming systems can be used as underlying reasoning engine.
In particular, \PLP\ is conceived as a front-end to the logic programming 
systems \DLV{}
and \SM{}. 
\end{abstract}

\section{General Information}

Several approaches have been introduced in recent years undertaking the
ability to express preference information within declarative knowledge 
representation formalisms~\cite{baahol93a,brewka94a,brewka96a,gelson97a,zhafoo97a,breeit99a}. 
However, most of these methods treat preferences at the meta-level and 
require a change of the underlying semantics. For instance, incorporating 
explicit orderings of logic programming rules is realized by modifying 
the fixed-point conditions characterizing answer sets or well-founded 
consequences. 
As a result, implementations need in general fresh algorithms and 
cannot rely on existing systems computing the regular (unordered) formalisms. 

In this paper, we describe a compiler, \PLP, for preferred answer 
sets which evades the need of new algorithms. 
\PLP\ is based on an approach for expressing preference information 
\emph{within} the framework of standard answer set semantics.
The general technique is described in \cite{delschtom:nmr00} and 
derives from a methodology for addressing preferences in default 
logic first proposed in \cite{delsch97a}.
We begin with an {\em ordered} logic program, which is an extended
logic program in which rules are named by unique terms and in which
preferences among rules are given by a new set of atoms of the form
$s \prec t$, where $s$ and $t$ are names.
Such an ordered logic program is then transformed into a second, regular,
extended logic program wherein the preferences are respected, in the sense
that the answer sets obtained in the transformed theory correspond to
the preferred answer sets of the original theory.
The approach is sufficiently general to allow the specification of 
preferences among preferences, preferences holding in a particular 
context, and preferences holding by default.

\PLP\ has been implemented in Prolog and serves as a front-end for 
the logic programming systems
\texttt{dlv}~\cite{dlv97a} and \texttt{smodels}~\cite{niesim97a};
it has been developed under the \ECLiPSe{} Constraint Logic Programming 
System and comprises roughly 300 lines of code. It runs under any 
operating systems for which \DLV\ or \SM\ are executable, provided 
\ECLiPSe\ Prolog is available. 
Apart from the pretty-printer, we only employ standard Prolog 
programming constructs.

\section{Description of the System}

The logic programming systems \DLV\ and \SM\ represent state-of-the-art 
implementations of the stable model semantics~\cite{gellif88b}, the 
former system also admitting strong negation and disjunctive rules. 
As well, several front-ends for these systems have been developed with 
the purpose of providing direct encodings of advanced reasoning problems, 
like, \egc diagnostic reasoning or planning~\cite{eite-etal-98n,eite-etal-00}.

\PLP\ realizes a further front-end for these systems, handling ordered 
logic programs.
The structure of \PLP\ is divided into five files:
\begin{itemize}
\item The actual compiler is contained in the file \texttt{pref.pl} (or
  \texttt{pref}$x$\texttt{.pl}, where $x$ is the actual version number).

\item The file \texttt{grounder.pl} takes care of programs with variables and
  non-atomic term structures.

\item
  The file \texttt{pp.pl} provides a pretty-printer for
  the files resulting from the compilation.
  (This module relies on features of \ECLiPSe{} for transforming Prolog's
  variables like \texttt{\_4711} into more readable ones like \texttt{X}.
  This is necessary in view of processing the files with \DLV{} and \SM{}.)
\item The files \texttt{dlv.pl} and \texttt{smodels.pl} contain code specific to
  the respective logic programming systems.
\end{itemize}

The general function of \PLP\ works in three stages:

\begin{enumerate}

\item loading the compiler (which charges also the Prolog files 
\texttt{pp.pl}, \texttt{grounder.pl},
\texttt{dlv.pl}, and \texttt{smodels.pl});

\item compiling the original Prolog file
(the compiler is pretty verbose and displays also intermediate versions 
of the compiled program);

\item call of \DLV{} or \SM.
\end{enumerate}

The present version of \PLP\ is written for the following releases of \DLV{} and \SM:

\begin{description}
\item[$\DLV:$] release from November 24, 1999.
\item[$\SM:$] version 1.12, including \texttt{parse-1.3}.
Adaptions for \texttt{lparse} and \texttt{pparse} (and thus
  \SM\texttt{-2}) are under development.
\end{description}

The current prototype of \PLP\ can be found at the following URL:
\begin{quote}
\texttt{http://www.cs.uni-potsdam.de/\~{}torsten/plp/}.
\end{quote}

\section{Applying the System}

\subsection{Methodology}
                                        

As part of the general logic programming paradigm, ordered logic programs
deal with knowledge representation tasks in a \emph{declarative} way. 
As well, the methodology of representing problems using ordered logic programs 
is essentially the same as the methodology of representing problems in terms of 
standard logic programs. However, the admission of explicit preference relations
within an ordered logic programs offers a convenient way to represent problems 
which would otherwise be somewhat tedious to handle.

\subsection{Specifics}
                                        

\PLP\ offers several methods how preferences can be encoded. To begin with, 
preferences between single \emph{rules} can be specified. As well, programs 
can contain preference relations between \emph{sets} of (names of) rules. 
Finally, in addition to the general approach discussed in \cite{delschtom:nmr00}, 
our compiler can also handle rules with variables.

\paragraph{Preferences Between Single Rules.} 
The specification of preference relations between single rules 
is achieved by associating names to certain (or all) rules of the given 
program, and by using atoms of the form $\PREC{n}{n'}$, where $n,n'$ 
are names of rules, to express a preference relation between rules 
named $n$ and $n'$, respectively. 
Informally, atom $\PREC{n}{n'}$ states that the rule named $n$ has 
higher priority than the rule named $n'$. Preference atoms may occur 
in any program rules, thus defining preference information in 
a \emph{dynamic way}. 
For instance, the rule
\[
\PREC{n}{n'}\LPif p, \naf q
\]
expresses preference of rule named $n'$ over rule named $n$ in 
case that $p$ is known and $q$ is not known. 
Observe that $p$ and $q$ may themselves be preference atoms, 
or rely on other preference atoms.

Let us illustrate the use of preference information between 
single rules by the following program $\Pi$, representing 
the well-known penguin-birds-example, but where an additional preference
between two conflicting rules is specified:
\[
\begin{array}{lrcl}
  &penguin(tweety) &\leftarrow&
  \\[1ex]
  &bird(tweety) &\leftarrow&
  \\[1ex]
  r_1:&flies(tweety) &\leftarrow& bird(tweety), \\
  &&&not\ \neg flies(tweety)
  \\[1ex]
  r_2:&\neg flies(tweety) &\leftarrow& penguin(tweety), \\
  &&&not\  flies(tweety)
  \\[1ex]
      &n_1\prec n_2 &\leftarrow&
\end{array}
\]
Rules $r_1$ and $r_2$ are associated with names $n_1$ and $n_2$, which are
constants in the object language. 
It is not necessary to name each rule in a program, but only those 
which occur as an argument in a preference atom.
Fact
\(
n_1\prec n_2 \leftarrow
\)
expresses that $r_2$ has higher priority than $r_1$. 
Without this preference information, this program has two answer sets,
one containing $flies(tweety)$ and the other containing $\neg flies(tweety)$.

The above program $\Pi$ is expressed in our syntax as follows:
{\small\begin{tabbing}
\texttt{penguin(tweety).}\\
\texttt{bird(tweety).}\\[1ex]
\texttt{flies(tweety) :- } \= \texttt{name(1),}\\
                           \> \texttt{not neg flies(tweety),}\\ 
                           \> \texttt{bird(tweety).}
                              \\[1ex]
\texttt{neg flies(tweety) :- } \= \texttt{name(2),}\\ 
                               \> \texttt{not flies(tweety),}\\ 
                               \> \texttt{penguin(tweety).}
                                  \\[1ex]
\texttt{1 < 2.}
\end{tabbing}}
The atoms \texttt{name(1)} and \texttt{name(2)} are used to refer to the 
names of $r_1$ and $r_2$, respectively. 
Compiling the file yields the following result:

{\small\begin{tabbing}
\texttt{penguin(tweety).} \\
\texttt{bird(tweety).} \\[1ex]
\texttt{flies(tweety) :- ap(1).} \\[1ex]
\texttt{ap(1) :- } \=  \texttt{ok(1), not not}\verb+_+\texttt{flies(tweety),}\\ 
                   \>  \texttt{bird(tweety).}
                       \\[1ex]
\texttt{ap(2) :- } \>  \texttt{ok(2), 
                         not flies(tweety),}\\ 
                   \>  \texttt{penguin(tweety).}
                       \\[1ex]
\texttt{ok(N) :- } \>  \texttt{name(N), oko(N, 1), oko(N, 2).}
                       \\[1ex]
\texttt{bl(1) :- } \>  \texttt{ok(1), not}\verb+_+\texttt{flies(tweety).}
                       \\
\texttt{bl(1) :- } \>  \texttt{ok(1), not bird(tweety).}
                       \\
\texttt{bl(2) :- } \>  \texttt{ok(2), flies(tweety).}
                       \\
\texttt{bl(2) :- } \>  \texttt{ok(2), not penguin(tweety).}
                       \\[1ex]
\texttt{not}\verb+_+\texttt{flies(tweety) :-  ap(2).}
                       \\[1ex]
\texttt{prec(1, 2).}\\[1ex]
\texttt{oko(N, N) :- } \= \texttt{name(N).}
                          \\[1ex]
\texttt{oko(N, M) :- } \> \texttt{name(N), name(M),}\\
                       \> \texttt{not prec(N, M).} 
                          \\[1ex]
\texttt{oko(N, M) :- } \> \texttt{name(N), name(M),}\\
                       \> \texttt{prec(N, M), ap(M).}
                          \\[1ex]
\texttt{oko(N, M) :- } \= \texttt{name(N), name(M),}\\
                       \> \texttt{prec(N, M), bl(M).}
                       \\[1ex]
\texttt{name(2).}\\
\texttt{name(1).}
\end{tabbing}}
What is the meaning of the newly introduced predicates and rules? 
The predicates \texttt{ap/1}, \texttt{bl/1}, \texttt{ok/1}, and 
\texttt{oko/2} control the order of the applications of the original 
rules and guarantee that the intended preference relation is respected. 
Informally, \texttt{ap(n)} and  \texttt{bl(n)} express that the rule 
named \texttt{n} is already known to be ``applied'' or ``blocked'', 
respectively, whilst \texttt{ok(n)} states that it is acceptable to 
apply the rule associated with name \texttt{n}. 
Atom \texttt{ok(n)} is accepted on the basis of the auxiliary 
predicates \texttt{oko/2}.  
As well, the order relation \texttt{<} is captured by the predicate 
\texttt{prec/2}.

Piping the result into \DLV{} yields the following answer set:
{\small\begin{verbatim}
{true, name(1), name(2), penguin(tweety),
 bird(tweety), ok(2), oko(1,1), oko(2,1), 
 oko(2,2), prec(1,2), neg_prec(2,1), 
 ap(2), neg_flies(tweety), oko(1,2), 
 ok(1), bl(1)}
\end{verbatim}}
We obtain a single answer set that corresponds to the second 
answer set for the example without preferences. 
Also, it includes the additional atoms that stem from the 
necessary preference-handling rules, as discussed above.
If the user desires, these special-purpose atoms may be suppressed
from the output by using the option ``\texttt{nice}'' in the execution command.

\paragraph{Dynamic Preferences with Variables.}
\PLP\ can also deal with parametrized names. The use of variables 
admits the specification of conditional, context-sensitive 
preferences. Consider the following example:
{\small\begin{tabbing}
\texttt{bird(tweety). penguin(tweety). water}\verb+_+\texttt{shy(tweety).}\\
\texttt{bird(opus). emu(opus).}\\
\texttt{bird(scully). toy(scully).}\\[1ex]

\texttt{\hphantom{neg }flies(X) :- } \= \texttt{name(r1(X)), not neg flies(X),}\\ 
                                     \> \texttt{bird(X).}
                                     \\[1ex]
\texttt{neg flies(X) :- } \> \texttt{name(r2(X)), not     flies(X),}\\
                          \> \texttt{penguin(X).}
                          \\[1ex]
\texttt{neg flies(X) :- } \> \texttt{name(r3(X)), not     flies(X), emu(X).}
                          \\[1ex]
\texttt{neg flies(X) :- } \> \texttt{name(r4(X)), not     flies(X), toy(X).}
                          \\[1ex]
\texttt{(r1(X) < r2(X)) :- not water}\verb+_+\texttt{shy(X).}\\[1ex]
\texttt{(r1(X) < r3(X)).}
\end{tabbing}}
For treating such programs, they are preprocessed in two steps:
\begin{enumerate}
\item rules with variables are replaced by their ground instances;
\item names with a complex term structure are turned into constants
  (eg.\ \texttt{name(r(f(c)))} is turned into \texttt{name(r\_f\_c)}).
\end{enumerate}

The above program yields the following answer sets:

{\small\begin{tabbing}
\verb+{+\texttt{bird(tweety), bird(opus), bird(scully),}\\
\texttt{ penguin(tweety), water}\verb+_+\texttt{shy(tweety),}\\
\texttt{ emu(opus), toy(scully), flies(tweety),}\\
\texttt{ flies(scully), neg}\verb+_+\texttt{flies(opus)}\verb+}+
\\[1ex]
\verb+{+\texttt{bird(tweety), bird(opus), bird(scully),}\\
\texttt{ penguin(tweety), water}\verb+_+\texttt{shy(tweety),}\\
\texttt{ emu(opus), toy(scully), flies(scully),}\\
\texttt{ neg}\verb+_+\texttt{flies(tweety), neg}\verb+_+\texttt{flies(opus)}\verb+}+
\\[1ex]
\verb+{+\texttt{bird(tweety), bird(opus), bird(scully),}\\
\texttt{ penguin(tweety), water}\verb+_+\texttt{shy(tweety),}\\
\texttt{ emu(opus), toy(scully), flies(tweety),}\\
\texttt{ neg}\verb+_+\texttt{flies(opus), neg}\verb+_+\texttt{flies(scully)}\verb+}+
\\[1ex]
\verb+{+\texttt{bird(tweety), bird(opus), bird(scully),}\\
\texttt{ penguin(tweety), water}\verb+_+\texttt{shy(tweety),}\\
\texttt{ emu(opus), toy(scully), neg}\verb+_+\texttt{flies(tweety),}\\
\texttt{ neg}\verb+_+\texttt{flies(opus), neg}\verb+_+\texttt{flies(scully)}\verb+}+
\end{tabbing}}
\noindent
By using the option ``\texttt{nice}'', we suppressed the special-purpose atoms in the displayed output.

\paragraph{Set-ordered Programs.}
Generalizing the idea of preferences between rules, 
\emph{set-ordered programs} admit the specification of preference 
information between sets of rules.
Informally, if set $M'$ is preferred over set $M$, then $M$ 
is considered after {\em all\/} rules in $M'$ are found to be 
applicable, or some rule in $M'$ is found to be inapplicable. 

In order to express preferences between sets of rules, set-ordered 
programs associate unique names to sets of rules and express 
preference information in terms of atoms $\PREC{m}{m'}$, where 
$m,m'$ are names of sets of rules. 
Following our previous convention, $\PREC{m}{m'}$ states that 
the set named $m'$ is preferred over the set named $m$.

The idea behind sets of preferences is illustrated by the following 
example dealing with ``car features'':
\begin{quote}
  Consider where in buying a car one ranks the price (``\texttt{expensive}'') over
  safety features (``\texttt{safe}'') over power (``\texttt{powerful}''), but safety
  features together with power is ranked over price.
\end{quote}
This is expressed by the following set-ordered program:
{\small\begin{tabbing}
\texttt{expensive} \= \texttt{:- name(e), not neg expensive.}\\
\texttt{powerful}  \> \texttt{:- name(p), not neg powerful.}\\
\texttt{safe}      \> \texttt{:- name(s), not neg safe.}\\[1ex]
\texttt{neg expensive} \= \texttt{:- \hphantom{expensive,} powerful, safe.}\\
\texttt{neg powerful}  \> \texttt{:- expensive, \hphantom{powerful,} safe.}\\
\texttt{neg safe}      \> \texttt{:- expensive, powerful\hphantom{, safe}.}\\[1ex]
\texttt{m1 : [p].}\\
\texttt{m2 : [s].}\\
\texttt{m3 : [e].}\\
\texttt{m4 : [p,s].}\\[1ex]
\texttt{m1 < m2.}\\
\texttt{m2 < m3.}\\
\texttt{m3 < m4.}
\end{tabbing}}\noindent
Here, \texttt{m1} is the name of the set \texttt{[p]} comprising 
the rule named \texttt{p} (and analogously for \texttt{m2}, 
\texttt{m3}, and \texttt{m4}). 
More generally, sets are expressed in an extensional way by:
{\small\begin{verbatim}
<set-name> : [<rule-name>,...,<rule-name>] .
\end{verbatim}}\noindent
As well, preferences on individual rules are expressed via 
the corresponding singleton sets.

Engaging our compiler and piping the described program into 
\DLV, we obtain the following answer set:
{\small\begin{verbatim}
{powerful, safe, neg_expensive}
\end{verbatim}}
\noindent
Again, in the displayed result we suppressed the special-purpose 
atoms by engaging the option ``\texttt{nice}''.

\subsection{Users and Usability} 


As for any logic programming system, proper usage of \PLP\ requires 
the user's ability to formalize a given problem adequately in terms
of an ordered (or set-ordered) logic program. 
However, as already pointed out above, the added benefit of ordered 
logic programs to allow the explicit specification of certain domain
knowledge in terms of preference declarations enriches the already 
versatile logic programming paradigm.

\section{Evaluating the System}

\subsection{Benchmarks}

Accepted benchmarks for nonmonotonic reasoning systems have been 
realized by the well-known TheoryBase system~\cite{chmamitr95a}. 
This test-bed provides encodings of various graph problems in terms 
of default theories and equivalent logic programs. 
Although these benchmark problems do not include ordered logic 
programs, they can nevertheless be used to evaluate \PLP.
The reason for this is given by the fact that our method reduces 
ordered logic programs to regular logic programs.
 Moreover, since ordered logic programs agree semantically with 
regular logic programs if no preference information is present, 
worst-case classes for regular logic programs are also worst-case 
classes for ordered logic programs.

\subsection{Comparison} 


Comparisons between different theorem provers are often difficult 
to realize because of incompatibilities of the respective underlying 
semantics. 
Even if two systems are based on the same formalism, they may 
represent different syntactical fragments rendering significant 
comparisons nigh impossible.

A method to address some of these shortcomings is to perform 
comparisons taking the representational power of the implemented 
formalisms into account.
That is to say,  one chooses some ``natural'' problems, encodes 
it with respect to the specific methodologies associated with the 
implemented formalisms, and uses the resultant instances as 
queries of the respective systems.
So, basically, different systems are compared on the basis of 
(possibly) different representations \emph{of the same problem}. 
In this sense, \PLP\ can be compared with other logic programming 
systems, whether or not they do support the explicit specification of
preference relations.
In fact, we may even conduct comparisons between \PLP\ and \DLV\ or \SM\
\emph{directly}, by representing a problem equivalently with and without
the use of ordering relations, and so being able to evaluate the representational power of explicit preferences.

\subsection{Problem Size} 


\DLV\ and \SM\ are state-of-the-art implementations successfully 
demonstrating their respective ability to process large and complex 
problem descriptions.
Because our compiler is an efficient translator (the resultant 
logic programs are polynomial in the size of the input program), 
\PLP\ likewise scales up to real-life problem representations.

\bibliographystyle{aaai}

\end{document}